\documentclass[a4paper, 10pt, conference]{ieeeconf}      
\pdfoutput=1
\IEEEoverridecommandlockouts                              

\usepackage{graphics} 
\usepackage{epsfig} 
\usepackage{mathptmx} 
\usepackage{times} 
\usepackage{amsmath} 
\usepackage{amssymb}  
\usepackage{cite}
\usepackage{amsmath,amssymb,amsfonts}
\usepackage{algorithmic}
\usepackage{graphicx}
\usepackage{caption}
\usepackage{subcaption}
\usepackage{textcomp}
\usepackage{xspace}
\usepackage{balance}
\usepackage{paralist}
\usepackage{url}
\usepackage[draft]{hyperref}
\usepackage[dvipsnames]{xcolor}
\usepackage{multirow}
\usepackage{subcaption}
\usepackage{tabularx}
\usepackage{multirow}
\usepackage{dsfont}

\title{\LARGE \bf
Fault-Tolerant Perception for Automated Driving A Lightweight Monitoring Approach
}

\author{
	Cornelius Buerkle$^{1}$, Florian Geissler$^{1}$, Michael Paulitsch$^{1}$, and Kay-Ulrich Scholl$^{1}$
	\thanks{$^{1}$Intel Labs, Germany. Emails: \url{Cornelius.Buerkle@intel.com},  \url{Florian.Geissler@intel.com},  \url{Michael.Paulitsch@intel.com}, \url{Kay-Ulrich.Scholl@intel.com}}%
}

\newcommand{\xstate}[1]{o_{#1}} 
\newcommand{\dt}{\delta t} 
\newcommand{\xt}{{x}_t } 
 
\newcommand{\yt}{{y}_t } 
 
\newcommand{\vt}{{v}_t } 
 
\newcommand{\tht}{{\theta}_t } 
 
\newcommand{\at}{{a}_t} 
 
\newcommand{\wt}{{\omega}_t }

\newcommand{\ate}{\hat{a}_t } 
\newcommand{\wte}{\hat{\omega}_t }

\newcommand{\xtpe}{\hat{x}_{t+\dt}} 
\newcommand{\ytpe}{\hat{y}_{t+\dt}} 

\newcommand{\thtm}{\tilde{\theta}_t} 
\newcommand{\thtpm}{\tilde{\theta}_{t+\dt}} 
\newcommand{\vtm}{\tilde{v}_t } 
\newcommand{\vtpm}{\tilde{v}_{t+\dt}} 
\newcommand{\xtm}{\tilde{x}_{t} } 
\newcommand{\ytm}{\tilde{y}_{t} } 
\newcommand{\xtpm}{\tilde{x}_{t+\dt}} 
 
\newcommand{\vm}{\tilde{v}} 
\newcommand{\dv}{\text{d}v} 
\newcommand{\dvmin}{\text{d}v_{\text{min}}} 
\newcommand{\xm}{\tilde{x}} 
\newcommand{\ym}{\tilde{y}} 
\newcommand{\thm}{\tilde{\theta}} 

\newcommand{\aacc}{a_{\text{acc}}} 
\newcommand{\abr}{a_{\text{br}}} 
\newcommand{\wmax}{\omega_{\text{max}}}

\newcommand{\fx}{\delta x} 
\newcommand{\fy}{\delta y} 

\newcommand{\dxmin}{\text{d}x_{\text{min}}}  
\newcommand{\mt}{\text{m}}  
\newcommand{\scd}{\text{s}}  
\newcommand{\mscd}{\text{ms}}  

\newcommand{\lidar}{\text{LiDAR}} 
\newcommand{\tautp}{\tau_{\text{tp}}}
\newcommand{\taufn}{\tau_{\text{fn}}}

\newcommand{\cgrid}{c_{\text{grid}}}
\newcommand{\dsigma}{\Delta B\left(\Sigma(o)\right)}
\newcommand{\gsens}{\gamma_{\text{sens}}}
\newcommand{\gplaus}{\gamma_{\text{plaus}}}

\newcommand{\treg}{${}^{\text{\textregistered}}$}
\newcommand{\ttm}{${}^{\text{\texttrademark}}$}

\begin{document}

\maketitle
\thispagestyle{empty}
\pagestyle{empty}

\begin{abstract}
While the most visible part of the safety verification process of automated vehicles concerns the planning and control system, it is often overlooked that safety of the latter crucially depends on the fault-tolerance of the preceding environment perception. 
Modern perception systems feature complex and often machine-learning-based components with various failure modes that can jeopardize the overall safety. At the same time, a verification by for example redundant execution is not always feasible due to resource constraints.
In this paper, we address the need for feasible and efficient perception monitors and propose a lightweight approach that helps to protect the integrity of the perception system while keeping the additional compute overhead minimal. 
In contrast to existing solutions, the monitor is realized by a well-balanced combination of sensor checks -- here using \lidar\ information -- and plausibility checks on the object motion history. It is designed to detect relevant errors in the distance and velocity of objects in the environment of the automated vehicle.
In conjunction with an appropriate planning system, such a monitor can help to make safe automated driving feasible.
\end{abstract}

\section{Introduction}

\begin{figure}
   \includegraphics[width=1\linewidth]{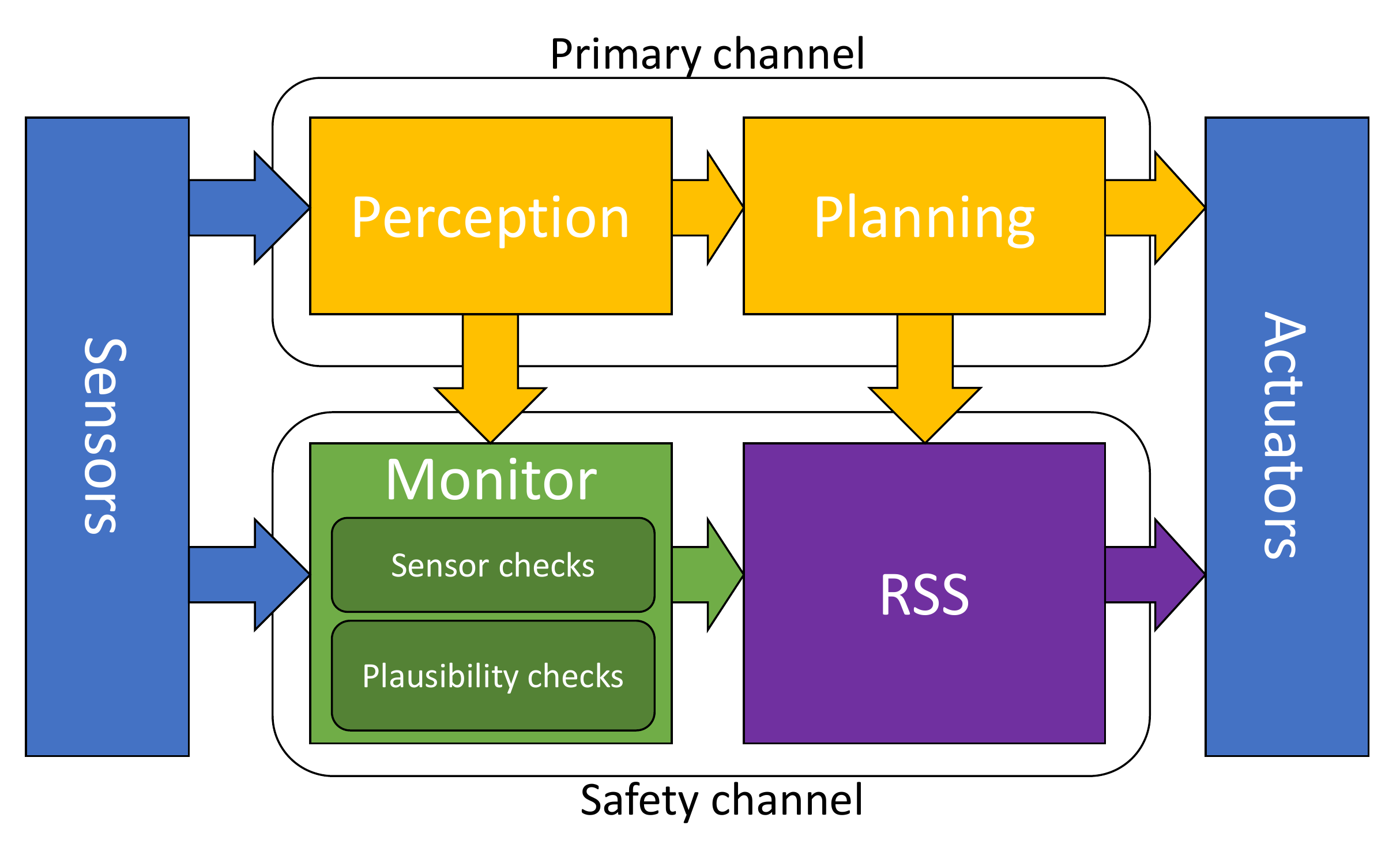}
   \caption{Proposed monitoring architecture}
\label{fig:Monitor}
\end{figure}

The development of automated vehicles (AVs) is one of the great technological challenges of today's society. While significant progress was made in the last years on the functional side, it remains yet unsolved how the safe operation of an AV can be assured. As illustrated in Fig.~\ref{fig:Monitor}, the AV system represents a complex, multi-stage compute stack involving most notably sensing, perception, planning and actuation, constituting the \textit{primary channel}. 
This compute chain is exposed to various sources of random faults, such as noise in the input data, hardware execution errors due to cosmic radiation, or systematic errors like software and hardware bugs, which can compromise the safe operation of the AV.  Particularly, this holds in the presence of artificial intelligence (AI)-based components, which are well-established in the perception and planning domain, but are for example highly affected by unbalanced or incomplete training data \cite{burton2017making}.

To achieve fault-tolerance of the primary channel, a secondary \textit{safety channel} is typically established (see Fig.~\ref{fig:Monitor}) that monitors the correct behaviour of the relevant primary functions.
In this work, we assume that solutions exist for the verification of the planning functions:
For example, Markov chains can be used to assess the safety of planned trajectories \cite{AlthoffSafety2007}. More recently, Responsibility Sensitive Safety (RSS) \cite{shalev2017formal} was proposed by Intel/Mobileye and it is currently adapted by the IEEE Standard P2846 \cite{ieeeAVSafety} for AV safety. 
All these approaches have in common that they rely on input provided by a perception system. 
\textit{Therefore, assuring the integrity of the perception system at reasonable costs and high efficiency is still a major challenge we address in this article.}

Various perception monitoring approaches have been tested:
One strategy is to create modular redundancies (e.g. duplex \cite{TeslaDMR2019}) or diverse function replication, e.g. leveraging different sensing modalities or algorithms \cite{safetyConceptsAV}. If the diversity is sufficiently large, it can be argued that this provides a certain level of protection against the aforementioned faults. 
It remains an open question though whether it is feasible to develop a sufficiently diverse system that achieves comparable quality on all channels: Instead, the channel with the lowest performance often dictates the overall system behaviour. Furthermore, the inherent challenges of learning-based systems remain and the computational cost is significantly increased. This results in the demand of accelerators and high-performance processors for which it is typically hard to ensure correctness and integrity as demanded by the functional safety standard ISO26262.      

In previous work \cite{buerkle2020onlineenv} we proposed a two-stage planning and perception monitoring architecture where the planning monitor was realized by RSS and the  perception monitor using a dynamic occupancy grid. While the detection performance of this implementation was very promising, the used dynamic occupancy grid relies on computational intense tracking algorithms. Therefore its compute requirements are a key challenge, unable to provide a lightweight solution that can run on FuSa-certified hardware.

To overcome this, we present here a simple, lightweight and low-cost, yet robust perception monitoring solution.
Our perception monitor consists of two components \textit{sensor checks} and \textit{plausibility checks}, see Fig.~\ref{fig:Monitor}. The former component performs a minimal processing of a high-confidence sensor source (here e.g. a \lidar) to accurately verify the positions and two-dimensional shapes of objects in a static occupancy grid. 
The latter component verifies the dynamics of objects based on a physical motion model -- thus not subject to sensing faults -- to identify for example unrealistic speeds, accelerations or discontinuous paths.
Both components run relatively simple functions, allowing for a light-weight implementation. Yet, we demonstrate that this monitoring concept can reliably detect speed and position errors that are large enough to be safety-critical in realistic situations.
The identified errors can be mitigated by the subsequent planning monitor using for example worst-case assumptions on the erroneous parameters. The focus of this current work is however on the detection, not mitigation of perception errors.

Importantly, the here described monitoring solution is generic and can be combined with any primary perception system and any planning module beyond RSS.

In summary, the article contributes the following novelties:
\begin{itemize}
\item We present a novel perception monitoring approach that leverages the strengths of both sensor and plausibility checking,
\item We demonstrate that this method provides a high precision and recall against the most relevant perception errors,
\item We verify that the implementation is low-cost and can be executed on certified hardware.
\end{itemize}

The remainder of this paper is structured as follow. Section~\ref{sec:relatedwork} will provide an overview of related work. Section~\ref{sec:realization} describes the envisioned monitor architecture in detail, followed by the experimental results in Section~\ref{sec:evaluation}. Finally, Section~\ref{sec:conclusion} concludes the paper.

\section{Related Work}
\label{sec:relatedwork}

General monitoring architectures and their trade-offs in terms of cost, safety, execution time etc. are for example discussed in \cite{safetyConceptsAV, aniculaesei2018toward}.
The authors of \cite{mehmed2020monitor, mehmed2020runtime} present the benefits of an inherently asymmetric commander-monitor concept compared to a triple modular redundancy system, and propose a run-time monitor that can assess the safety of a planned vehicle trajectory. An example for an asymmetric perception monitor based on dynamic occupancy grids is proposed in our previous work \cite{buerkle2020onlineenv}.
When machine-learning functions are utilized in AD systems, additional safety challenges emerge as discussed in \cite{burton2017making, koopman2018toward}.
Plausibility checking as a mean to improve perception at different stages of the sensor fusion process is explored for example in \cite{Versmold2006, Yavvari2018}. 
Use cases comprise the detection of implausible ghost vehicles \cite{Obst2015} or misbehaviour in vehicle-to-vehicle communication \cite{Ambrosin2019}.
In previous work, we have further harnessed plausibility checking to detect sensor faults in distributed sensor networks \cite{Geissler2020}.

\section{Realization}
\label{sec:realization}

\subsection{Errors}
As a conceptual limitation, a monitor of reduced complexity with respect to the primary channel, can detect only a subset of all possible perception errors of the latter. 
We thus restrict our analysis to the following errors which we consider as most relevant for the validation of the RSS perception input:
\begin{itemize}
\item \textbf{False negative detections} represent objects in the true environment that the AD system fails to detect. For example, an AI component may not be trained properly to detect a certain type of object, and thus miss it. These errors could result in unsafe planning and collisions.
\item \textbf{False positive detections} are observations of the AD system that do not exist in the true environment ("ghost" targets). These are not considered to be safety-critical, nevertheless they can negatively impact the comfort of the system as they might lead to unnecessary braking or evasive manoeuvres. 
\item \textbf{Position errors} result in an object being perceived at an incorrect location, potentially leading to unsafe manoeuvres or collisions. Depending on the object association, large position errors can be interpreted as a simultaneous false positive and false negative detection.
\item \textbf{Velocity errors} have the consequence that the motion of a perceived object is predicted in an incorrect way and might lead to false assumptions on required safety distances.
\end{itemize}
Our goal is to design a simplified monitor with sufficient detection capabilities to reliably identify the above errors, as quantified by the precision and recall metrics. A reduced precision of the monitor means increased false alarms, which directly affects the availability of the system. On the other hand, a reduced recall results in errors being missed by the monitor. Finding an optimal balance is a design challenge between the sensitivity and robustness of the architecture, especially as the input to the AV system is exposed to noise.
The verification of false positive and false negative detections requires a redundant sensor source, and is therefore handled by the occupancy grid of the sensor checks. We have studied this aspect in previous work \cite{buerkle2020onlineenv}. In this article, we focus on the ability of our perception monitor to detect critical position and speed errors using combined sensor and plausibility checks, see Section~\ref{sec:evaluation}. Those errors can be both transient or permanent.

Importantly, to properly calculate a safety envelope according to RSS definitions, only safety-critical manifestations of the mentioned error types need to be detected. Since the distinction of safe and unsafe errors depends generally on the explicit situation in a complex environment, it is not possible to define a universal threshold for, e.g., critical speed and position errors. 
We may however highlight two error magnitudes that typically lead to safety violations and serve as a benchmark for our monitor: For position errors, deviations of at least half the width of a lane, $\dxmin\approx 1\mt$, cause critical lane misassociations, or mistakes about pedestrians being on or off the road. Further, speed errors of about $\dvmin \approx 3\mt/\scd$ represent a typical difference between static and mobile pedestrians, and may also result in the misclassification of an object class, if typical speed estimates are partially or fully utilized for that purpose. 


\subsection{Sensor checks}

\subsubsection{Concept}
In order to realize the sensor checks we convert the \lidar\ point-cloud measurements in an occupancy grid. In contrast to the approach described in \cite{buerkle2020onlineenv}, we use for this work a classical occupancy grid \cite{Thrun2005}.
The occupancy grid provides a two-dimensional (2D) representation of the environment. Therefore, it divides the surrounding of the ego vehicle in cells of a given size, so each Cartesian position $\zeta$ within the grid is mapped to a specific cell. Each cell  contains the probability of the cell to be occupied by an obstacle. A point cloud can be easily transformed in such a grid representation, by projecting all points, that are measurements of obstacles, to the cells corresponding to their Cartesian position. Points that are part of the ground plane or of the environment that can be under-passed are excluded. Each point that falls within a cell increases the occupancy probability of that cell. Each grid cell therefore provides a spatial occupancy probability, for a given position, denoted $P(\zeta)$.  

In order to verify the object position we determine the region $A_o$ that is covered by an object $o$ at the current point in time, according to the primary input. This region is determined by the object state and its covariance matrix $\Sigma(o)$. We enlarge this region by an additional safety margin $\delta_{safe}$ to compensate for noise in the \lidar\ measurements and other undetected uncertainties in the system, see Fig.~\ref{fig:GridCapabilities}.
The coverage function $c(\zeta,o)$ that will determine the coverage for a position $\zeta$ and a given object $o$ takes the form
\begin{eqnarray}
c(\zeta,o) =
\left\{
\begin{array}{@{}l@{\thinspace}l}
1  & \text{ , if } \zeta \in A_o, \\
0  & \text{ , if } \zeta \notin A_o. \\
\end{array}
\right.
\end{eqnarray}

We then define metrics for the consistency, $\eta$, of the grid with a given object, $o$, and the conflict $\kappa$ of a given grid position with the overall environment as
\begin{align}
\label {consistency}
\eta(o) &= \max \limits_{ \zeta \in \ \mathds{R}^2}  ( c(\zeta,o) P(\zeta) ), \\
\label {conflict}
\kappa(\zeta) &=\min  \limits_{ \forall o \in O}  ( 1 - c(\zeta,o)) P(\zeta).
\end{align}
This allows us to evaluate false detections in the primary channel. Using two decision thresholds $\tautp$ and $\taufn$ and boolean mapping functions, denote $f, f'$, we have
\begin{eqnarray}
f(o) =
\left\{
\begin{array}{@{}l@{\thinspace}l}
\text{false positive}  & \text{ , if } \eta(o) < \tautp, \\
\text{true positive}  & \text{ , if } \eta(o) \geq \tautp, \\
\end{array}
\right.
\end{eqnarray}
and
\begin{eqnarray}
f'(\zeta) = \left\{
\begin{array}{@{}l@{\thinspace}l}
\text{false negative}  & \text{ , if } \kappa(\zeta) > \taufn, \\
\text{no-conflict cell}  & \text{ , if } \kappa(\zeta) \leq \taufn. \\
\end{array}
\right.
\end{eqnarray}

With the functions $f, f'$, it's possible to determine whether there are false positive or false negative detections. 

\subsubsection{Theoretical analysis of position errors}
A position error can be seen as a combination of a false positive and a false negative detection, as described above. 
The capability of the sensor check monitor component to detect position errors is determined by the grid size $\cgrid$, determining the spatial resolution, the expected noise $\Sigma(o)$ of the sensor input, and a safety margin denoted $\delta_{safe}$.
Robustness against noise and other non-faulty deviations of the sensor and object data is needed to assure a low false alarm rate.
\begin{figure}[h]
\centering
\includegraphics[width=0.5\textwidth]{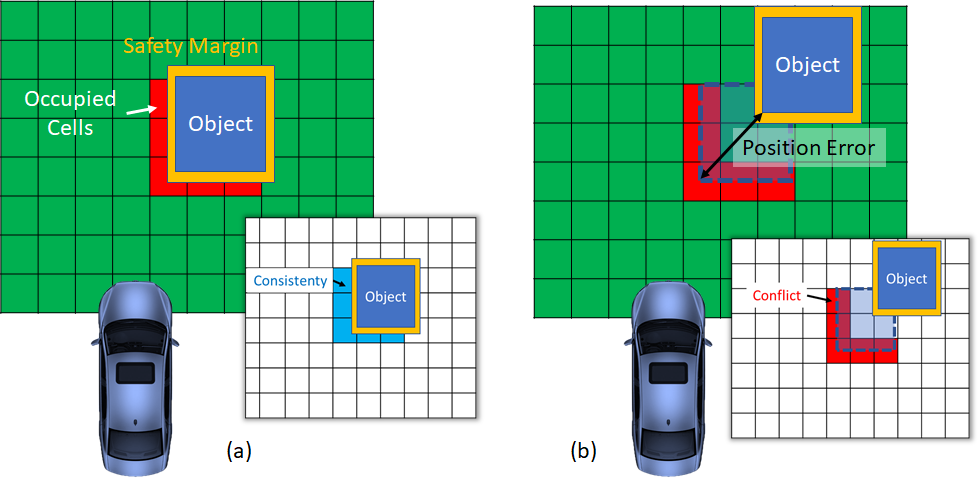}
\caption{Illustration of a position error detection with sensor checks.}
\label{fig:GridCapabilities}
\end{figure}

As visualized in Fig.~\ref{fig:GridCapabilities}, the grid is occupied around the part of a given object's bounding box that is hit by the \lidar\ beams, forming typically a "L"-shape. 
A position error can be detected when the conflict of grid cells within $A_{o}$ exceeds the threshold $\taufn$. It should be noted, that depending on the original location of individual measurement points within a grid cell, a given position error may or may not be sufficient to shift the measurement from one grid cell to another, and hence produce a conflict or not. 
The minimum position error $\dxmin$, that is \textit{guaranteed detectable}, can be estimated as the largest intra-cell shift diagonal to the grid orientation, 
\begin{equation}
\dxmin = \sqrt{2}(\cgrid + \delta_{safe}) + \gsens \dsigma
\label{eq:gridCapabilities}
\end{equation}
Here, $\dsigma$ represents the measurement uncertainty of the object's border, which can be approximated from the uncertainty of position $x,y$ and dimensions $L,W$ (length and width) as
$\dsigma = ||\Delta x + \Delta L, \Delta y + \Delta W||$. Those  margins $\Delta x$ etc. are obtained directly from the covariance matrix $\Sigma(o)$, and $|| \ldots ||$ is the Euclidean norm. Further, the factor of $\gsens$ in Eq.~\ref{eq:gridCapabilities} represents a specific confidence that the true object border is within the respective error margins and can be fine-tuned to control the sensitivity of the sensor checks. 
In practice, depending on object orientation and grid alignment, most position errors will fall below this worst-case estimate and are therefore detected already at smaller displacements, see Sec.~\ref{sec:evaluation}.





\begin{figure*}[htpb]
\centering
\begin{subfigure}{0.45\textwidth}
  \centering
	\includegraphics[width=1.\textwidth]{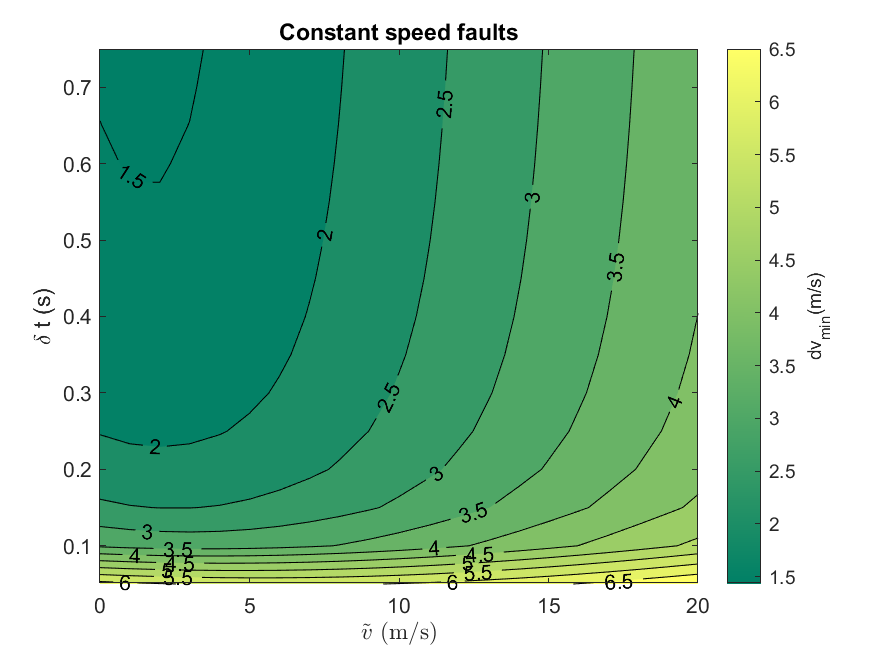}
	\caption{}
	\label{fig:dv_const}
\end{subfigure}
\begin{subfigure}{0.45\textwidth}
  \centering
	\includegraphics[width=1.\textwidth]{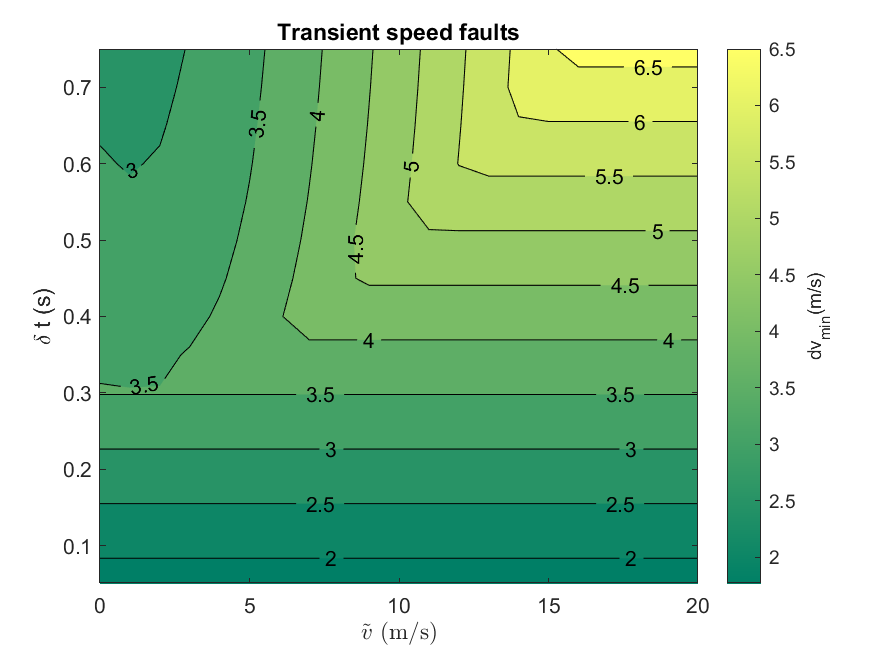}
	\caption{}
	\label{fig:dv_trans}
\end{subfigure}
\caption{Demonstration of the minimal speed error $\dvmin$ detectable by the plausibility history check, depending on interval time $\dt$ and object speed $\vm$. 
Parameters are chosen as $\Delta \vm = 1\mt/\scd$, $\Delta \xm = \Delta \ym = 0.1\mt$ and the values in Tab.~\ref{tab:sim_parameters}.}
\label{fig:dv_theo}
\end{figure*}

It is important to mention that with the proposed sensor checks only position errors that result in a \textit{larger distance} from the ego vehicle can be reliably detected: Position errors that cause the object to appear closer to the ego vehicle, on the other hand, are more difficult to detect as the object region $A_o$ might overlap with the real object edge. Nevertheless, we argue that errors of the latter type are typically not safety-critical as an erroneously perceived shorter distance will automatically lead to a more cautious behaviour of the vehicle.

\subsection{Plausibility checks}

\subsubsection{Motion model}
To analyze the object dynamics, we define a reduced object state at time $t$ by
\begin{equation}
\xstate{t} = \left(\xt, \yt, \vt, \tht, \at, \wt \right)^T,
\end{equation}
where $x$ and $y$ is a two-dimensional position, $v$ is the speed (absolute value of the velocity vector), $\theta$ the heading angle, $a$ the longitudinal acceleration, and $\omega$ the turn rate.
We adopt a constant turn rate and acceleration (CTRA) evolution during a time interval $\dt$, which takes the form \cite{Schubert2008}
\begin{equation}
\xstate{t+\dt} = \xstate{t} + 
\left(\begin{matrix}
	\fx(\dt, \vt, \tht, \at, \wt) \\
	\fy(\dt, \vt, \tht, \at, \wt) \\
	\wt \dt \\
	\at \dt \\
	0 \\
	0
\end{matrix}\right)
\end{equation}
Assuming sufficiently high sensor update rates for pseudo real-time modeling, we can focus on small time intervals and simplify the CTRA model by expanding around $\dt \approx 0$,
\begin{align}
\label{eq:mot_mod}
\begin{split}
& \fx = \vt \dt \cos(\tht) +\frac{\dt^2}{2}\left(\at \cos(\tht) - \vt \wt \sin(\tht) \right), \\
& \fy =  \vt \dt \sin(\tht) +\frac{\dt^2}{2}\left( \at \sin(\tht) + \vt \wt \cos(\tht) \right).
\end{split}
\end{align}

The plausibility check verifies whether or not the object displayed a plausible motion during the last interval $\dt$. 
Let us denote with $\tilde{x}$ variables that are measured, and with $\hat{x}$ variables that are predicted.
Then, in a first step, we estimate the non-observed variables turn rate and acceleration from speed and heading measurements
\begin{align}
\label{eq:pred1}
\begin{split}
\wte &= (\thtpm - \thtm)/\dt, \\
\ate &= (\vtpm - \vtm)/\dt.
\end{split}
\end{align}
Second, we use this estimate to predict the object position at time $t+\dt$ with the motion model of Eq.~(\ref{eq:mot_mod}),
\begin{align}
\label{eq:pred2}
\begin{split}
\xtpe &= \xtm + \fx(\dt, \vtm, \thtm, \ate, \wte), \\
\ytpe &= \ytm + \fy(\dt, \vtm, \thtm, \ate, \wte).
\end{split}
\end{align}
To evaluate the precision of our prediction, a standard error propagation \cite{Trivedi2016a} is performed assuming independence of the measured quantities $\xtm, \ytm, \vtm, \thtm$, and no error for $\dt$. 

Eventually, the motion of an object is considered implausible if at least one of the following conditions is true
\begin{align}
\label{eq:condition_w}
& \left(\wte - \Delta \wte > |\wmax| \right) \vee \left( \wte + \Delta \wte < -|\wmax| \right), \\
\label{eq:condition_a}
& \left(\ate - \Delta \ate > \aacc \right) \vee \left( \ate + \Delta \ate < \abr \right), \\
\label{eq:condition_xy}
& || \hat{\vec{x}}_{t} - \tilde{\vec{x}}_{t} || - \gplaus (||\Delta\hat{\vec{ x}}_{t}|| + ||\Delta\tilde{\vec{x}}_{t}||) > 0.
\end{align}
Here, $\Delta$ denotes the margin of error of a variable, and we have $\vec{x}_t=(x_t, y_t)$ as well as $\Delta \vec{x}_t=(\Delta x_t, \Delta y_t)$.
We introduce the parameter $\gplaus$ to control the sensitivity of the plausibility check, while $\wmax$, $\aacc$, $\abr$ are thresholds specifying a physically realistic maximum turn rate, forward acceleration, and brake acceleration, respectively.

\subsubsection{Theoretical analysis of speed errors}
\label{sec:plaus_theory}
To get a better intuition of the efficacy of the plausibility check, we perform a theoretical analysis of two different types of speed errors. 
As a minimal setup, we take an object moving with constant true speed, denote $\vm$, at two subsequent time steps with interval $\dt$, and inject a speed error of value $\dv$. 
We explore the impact of the time interval size and the object speed, while keeping for simplicity a constant heading $\thtm=\thtpm=0$ (and thus $\wte=0$ in this test scenario).
We calculate the minimal detectable speed error $\dvmin$, defined here as (positive or negative) speed error with the smallest absolute value, that can be detected by the plausible motion check.
Note that due to the position check of Eq.~(\ref{eq:condition_xy}) there will be a conceptual difference in detecting positive and negative speed errors in this model. Negative speed errors lead to slightly smaller predicted error margins for the position, compared to those of a positive speed error of the same magnitude, which can make them easier to detect.

\textbf{Permanent speed error:}

This error is represented by adding the constant shift $\vtm = \vtpm = \vm + \dv$ (importantly, this does not affect the next observed position $\xtpm$ which will evolve according to $\vm$ only). As the observed speeds are the same across all time steps, the estimated acceleration is zero and we can not reach the thresholds in Eq.~(\ref{eq:condition_a}).
The detection of speed errors is then based solely on the predicted position, which will overshoot or undershoot the next measured position. 
We can see two important trends governing $\dvmin$ in Fig.~\ref{fig:dv_theo}\subref{fig:dv_const}: 
An increase in $\vm$ leads to larger error margins for the predicted positions, given a nonzero uncertainty in the heading and turn rates. 
The minimal detectable speed error therefore grows with the object speed, except for a regime of small speeds $0 \leq \vm \lesssim \dvmin/2$, where the minimal speed errors is larger than two times the actual speed. 
An increase in the time interval $\dt$ both increases the prediction gap and the prediction error, where we find that typical measurement uncertainties balance this interplay in favor of the former, such that delayed updates typically help in detecting constant speed errors.

\textbf{Transient speed error:}

We simulate this behavior by enforcing a speed $\vtm = \vm$ and $\vtpm = \vm + \dv$.
For such transient speed errors, the regime of small time intervals is typically dominated by the acceleration check in Eq.~(\ref{eq:condition_a}), resulting in a detection threshold of $\dvmin = \min(\aacc, -\abr)\dt + \sqrt{2}\Delta \vm$.
The minimal detectable error is then independent of the object speed, increasing the sensibility for transient speed errors, see Fig.~\ref{fig:dv_theo}\subref{fig:dv_trans}.
With increasing time intervals, we see a superposition of the two conditions in Eq.~(\ref{eq:condition_a}) and Eq.~(\ref{eq:condition_xy}), as the position prediction check becomes 
more and more relevant. For large $\dt$, the minimal detected error will be determined by the position prediction only, and we observe the same trends as in Fig.~\ref{fig:dv_theo}\subref{fig:dv_const}.

\section {Evaluation}
\label{sec:evaluation}

\subsection {Experimental setup} 

\begin{figure}[h]
\centering

\begin{subfigure}[b]{0.45\textwidth}
\includegraphics[width=\textwidth]{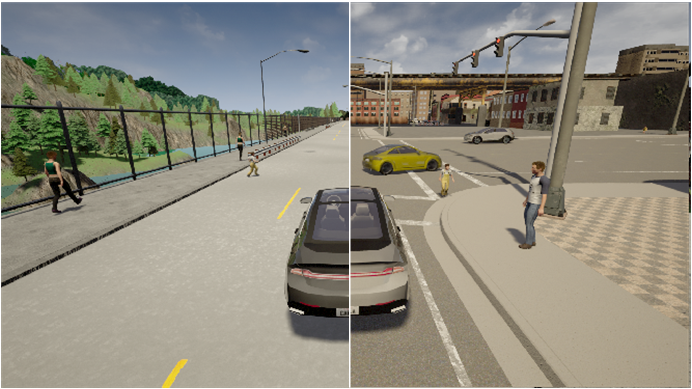}
\caption{Selected CARLA scenarios: (left) residential area with crossing pedestrians, (right) urban intersection with pedestrians and vehicles.}
\label{fig:Carla_int_ped}
\end{subfigure}

\vspace{.5cm} 
\begin{subfigure}[b]{0.45\textwidth}
\includegraphics[width=\textwidth]{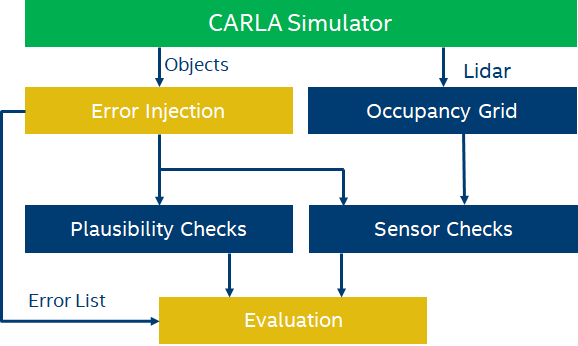}
\caption{Monitor evaluation process.}
\label{fig:Evaluation}
\end{subfigure}

\caption{System scenarios and process flow.}
\label{fig:setup}
\end{figure}

To evaluate our proposed safety approach, we use the CARLA simulator~\cite{Dosovitskiy17}, and equip an ego vehicle with a \lidar\ sensor.
The environment model is represented by the ground truth object information directly from CARLA. Subsequently, we inject position and speed faults into this ground truth object list and forward the manipulated object list to the monitor. For evaluation we compare the detected errors with the original error list to determine the efficacy of the implemented checks (Fig.~\ref{fig:Evaluation}). Importantly, the detection process for the plausibility checks comprises up to two subsequent time steps, since an error at a given time frame is typically detected by an implausible history at the next time frame.

For this paper, we have simulated two test scenarios in CARLA, see Fig.~\ref{fig:setup}. The first scenario represents a residential area with multiple spawned pedestrians, that randomly cross the street. The second scenario is an urban intersection featuring not only pedestrians but also cars and other vehicles. Those two setups were chosen to provide a diverse perception input to the ego vehicle, in terms of environment constellations and object types. In addition to that, we have tested position errors with the \textit{NuScenes} dataset \cite{nuscenes2019} as an example of non-simulated \lidar\ information. Each scenario duration was sufficiently long for the environment to contain more than $1000$ relevant object states in scope, in order to guarantee statistical significance.
All experiments use the parametrization of Tab.~\ref{tab:sim_parameters} unless stated otherwise.

\begin{center}
\captionof{table}{Simulation parameters}
\label{tab:sim_parameters}
\begin{tabular}{ c c } 
 \hline \hline
Name & value  \\ 
 \hline
 grid discretization & $100\mt \times 100 \mt$, $\cgrid = 0.5\mt$ \\
safety margin & $\delta_{safe} = 0.1\mt$ \\
 consistency/conflict bounds & $\tautp = 0.8$, $\taufn = 0.8$  \\ 
 max. accelerations & $\abr = -7\mt/\scd$, $\aacc = 7\mt/\scd$  \\ 
 Heading error & $\Delta \thm = 10^{\circ}$ \\
 max. turn rate & $\wmax = 90^{\circ}/0.2\scd$ \\
check sensitivities & $\gsens = 3$, $\gplaus = 1$ \\
 \hline \hline
\end{tabular}
\end{center}

\subsection {False alarm} 
An important design target of a monitor is a low false alarm rate, in the presence of noise, to maintain high system availability.

In order to evaluate the robustness of the proposed checks, we add Gaussian noise to the ground truth object positions, effectively increasing the measurement uncertainties $\Sigma (o)$.

\begin{figure}[h]
\centering
\includegraphics[clip,trim=0.cm 1.5cm 0.cm 1.5cm, width=0.5\textwidth]{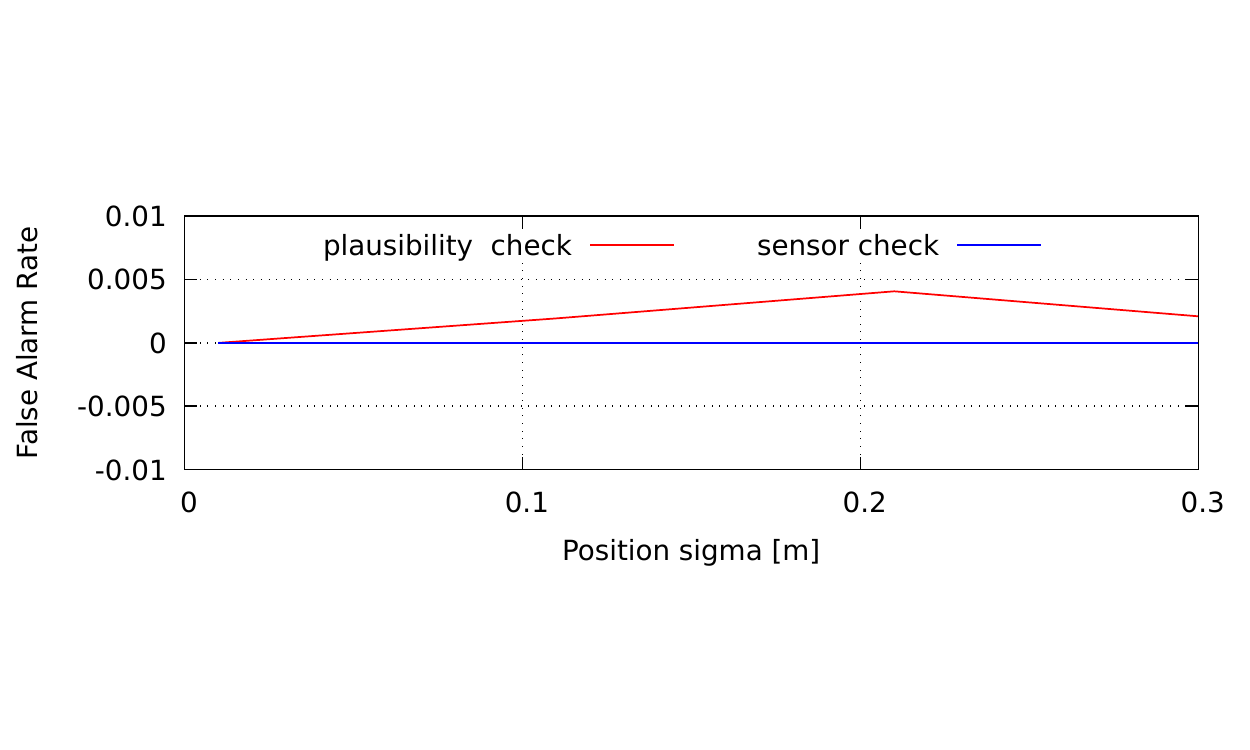}
\caption{False alarm rate for different position uncertainties.}
\label{fig:NoError}
\vspace{-0.5cm}
\end{figure}

For demonstration, we give results of the evaluation with the pedestrian scenario. Fig.~\ref{fig:NoError} shows that the sensor checks are very robust and produce zero false positives up to position noise of about $0.3\mt$. Similarly, for the plausibility checks, we find negligibly low false alarm rate remains below $5$\textperthousand.
This leads to a generally high precision of both monitor components in the experiments described in the next sections.

\subsection{Permanent position error} 
\label{sec:EvalPerPosError}
\begin{figure}[h]
\centering
\includegraphics [clip,trim=0.4cm 0.9cm 0.cm 1.1cm,width=0.5\textwidth]{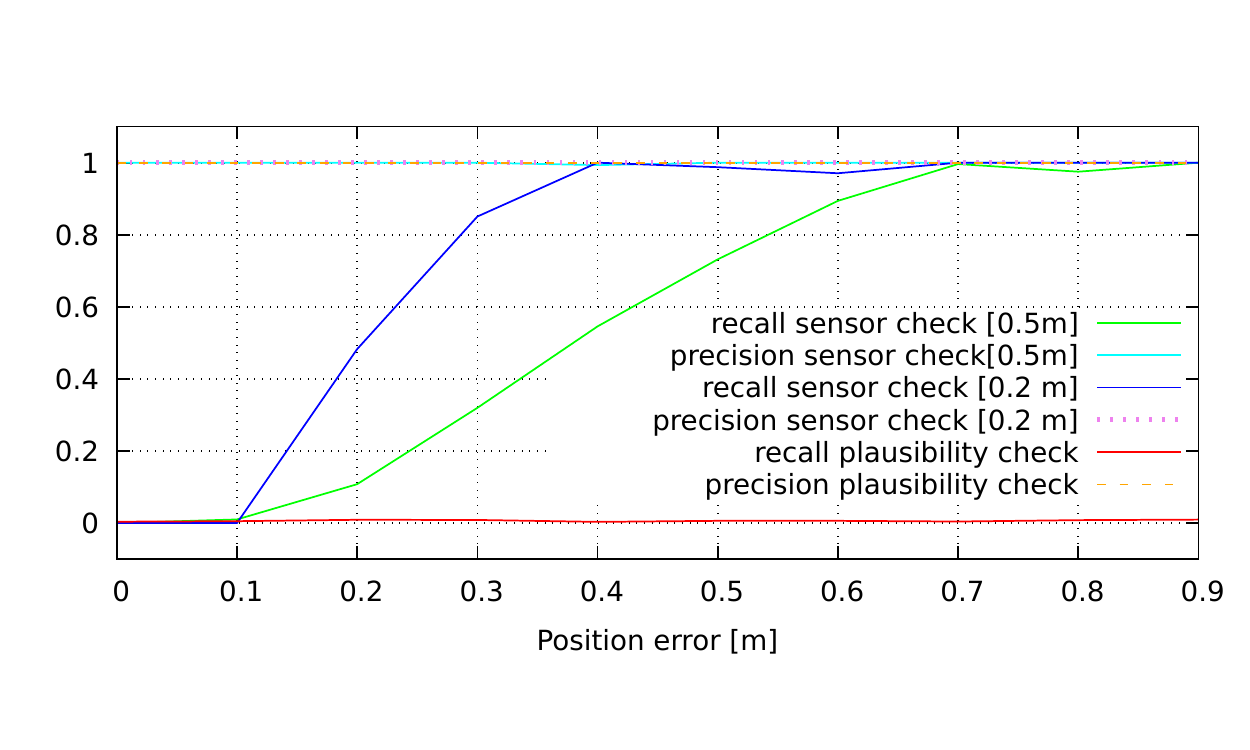}
\caption{Detection capabilities for permanent position errors. For the sensor checks, two different grid cell sizes  ($0.5\mt$ and $0.2\mt$) are tested.}
\label{fig:PositionError}
\vspace{-0.5cm}
\end{figure}

To analyse permanent position errors, we inject a constant position offset into the object list, increasing the distance of objects relative to the ego object. The evaluation results in Fig.~\ref{fig:PositionError}, using the pedestrian scenario, show that the sensor checks can reliably detect such errors if the offset is sufficiently large, which predominantly depends on the grid resolution. 

Explicitly, with a grid resolution of $0.2\mt$, position errors of a magnitude of $0.4 \mt$ can be reliable detected (recall $> 90 \%$), while with a cell size of $0.5\mt$ the minimum detectable error increases to $0.7\mt$.
The plausibility checks are not able to detect such permanent errors, as the consistency of the object history is not affected.

\begin{figure*}[h]
\centering
\begin{subfigure}[b]{\textwidth}
\centering
\includegraphics[width=0.5\textwidth]{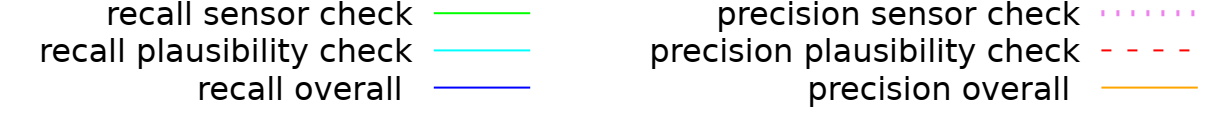}
\end{subfigure}
\caption*{\textbf{Pedestrian Scenario:}}
\begin{subfigure}[b]{0.32\textwidth}
\includegraphics[clip,trim=0.5cm 0.0cm 0.cm 0.0cm,width=\textwidth]{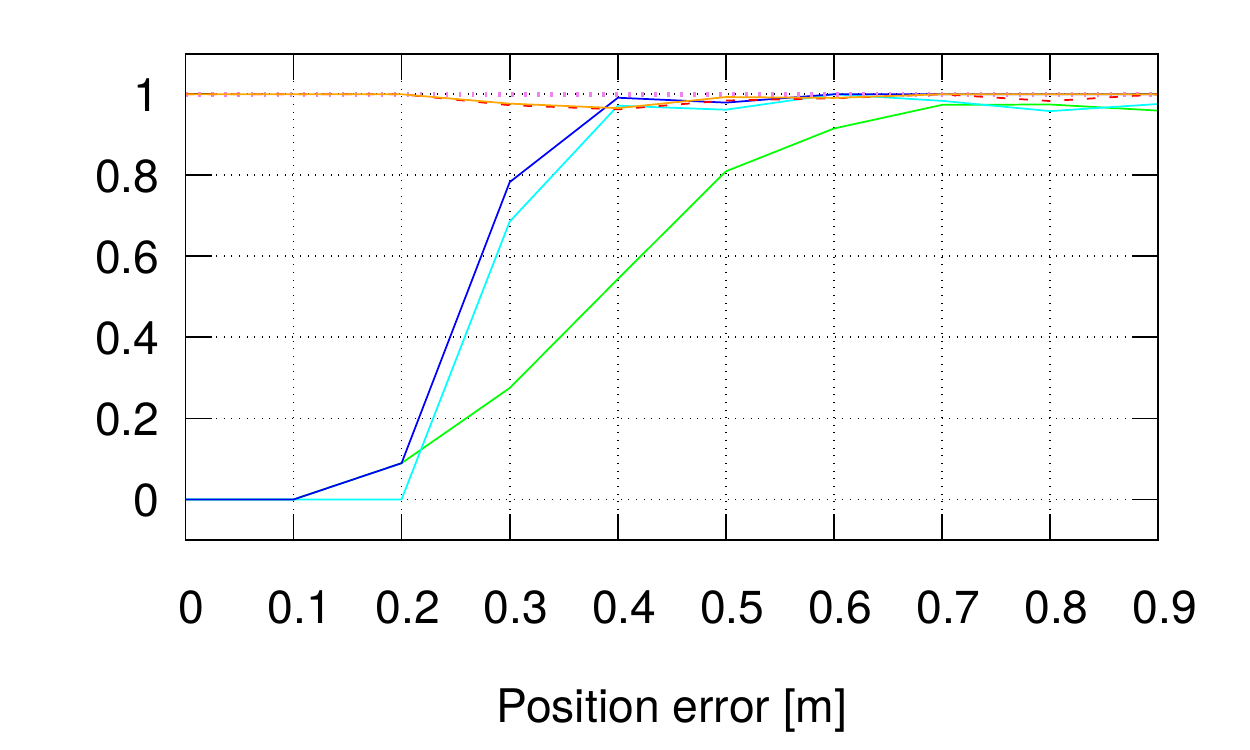}
\caption{Error rate 20\%}
\label{fig:PositionErrorRandom20}
\end{subfigure}
\begin{subfigure}[b]{0.32\textwidth}
\includegraphics[clip,trim=0.5cm 0.0cm 0.cm 0.0cm,width=\textwidth]{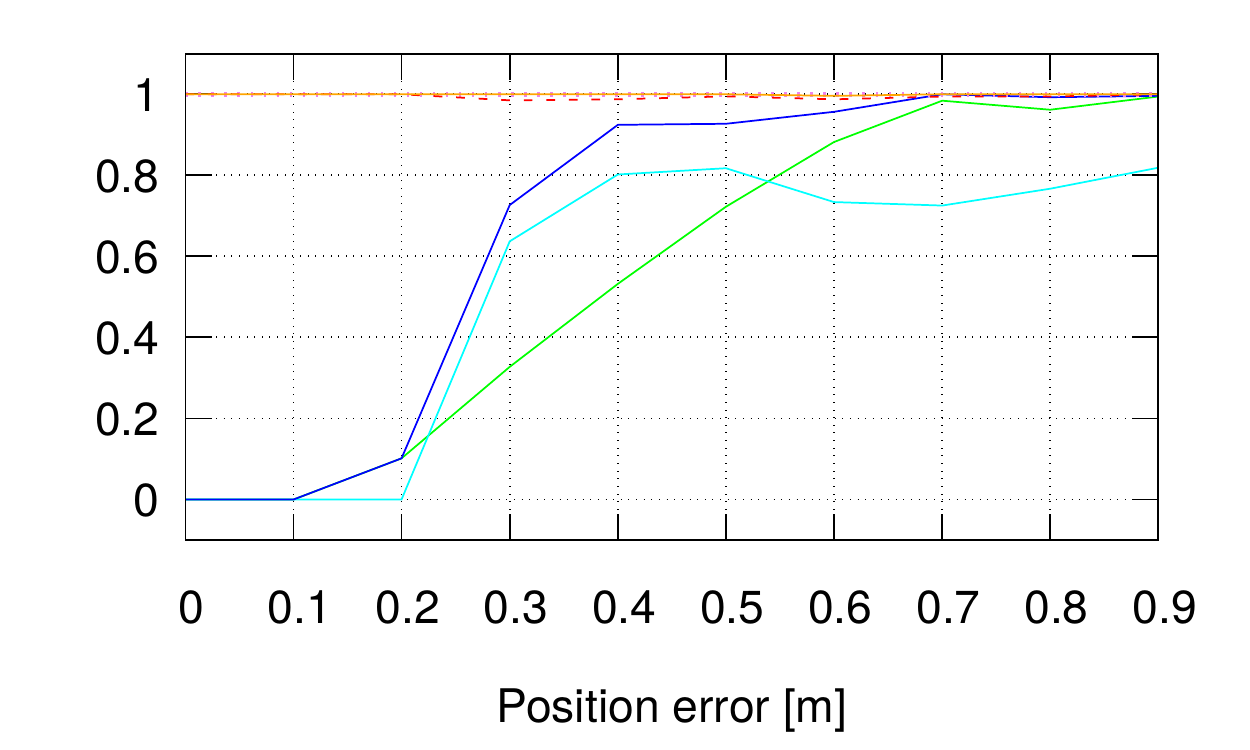}
\caption{Error rate 50\%}
\label{fig:PositionErrorRandom50}
\end{subfigure}
\begin{subfigure}[b]{0.32\textwidth}
\includegraphics[clip,trim=0.5cm 0.0cm 0.cm 0.0cm,width=\textwidth]{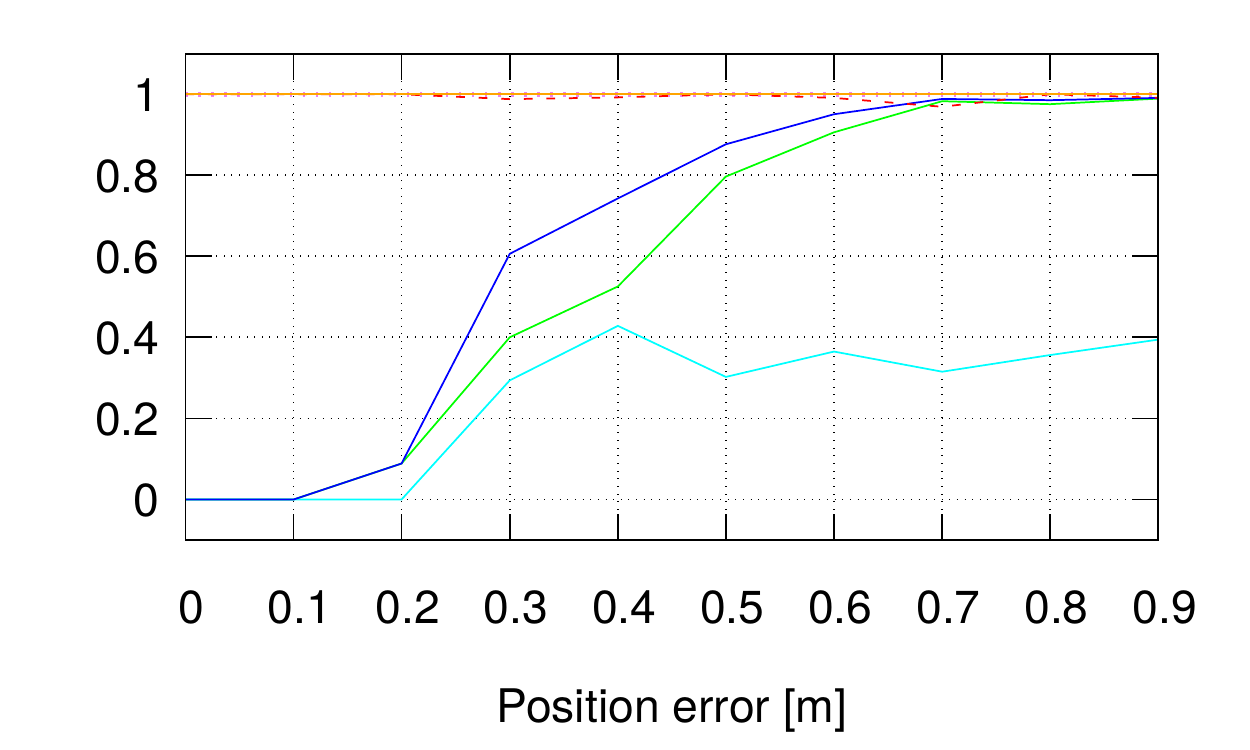}
\caption{Error rate 80\%}
\label{fig:PositionErrorRandom80}
\end{subfigure}

\caption*{\textbf{Intersection Scenario:}}
\begin{subfigure}[b]{0.32\textwidth}
\includegraphics[clip,trim=0.5cm 0.0cm 0.cm 0.0cm,width=\textwidth]{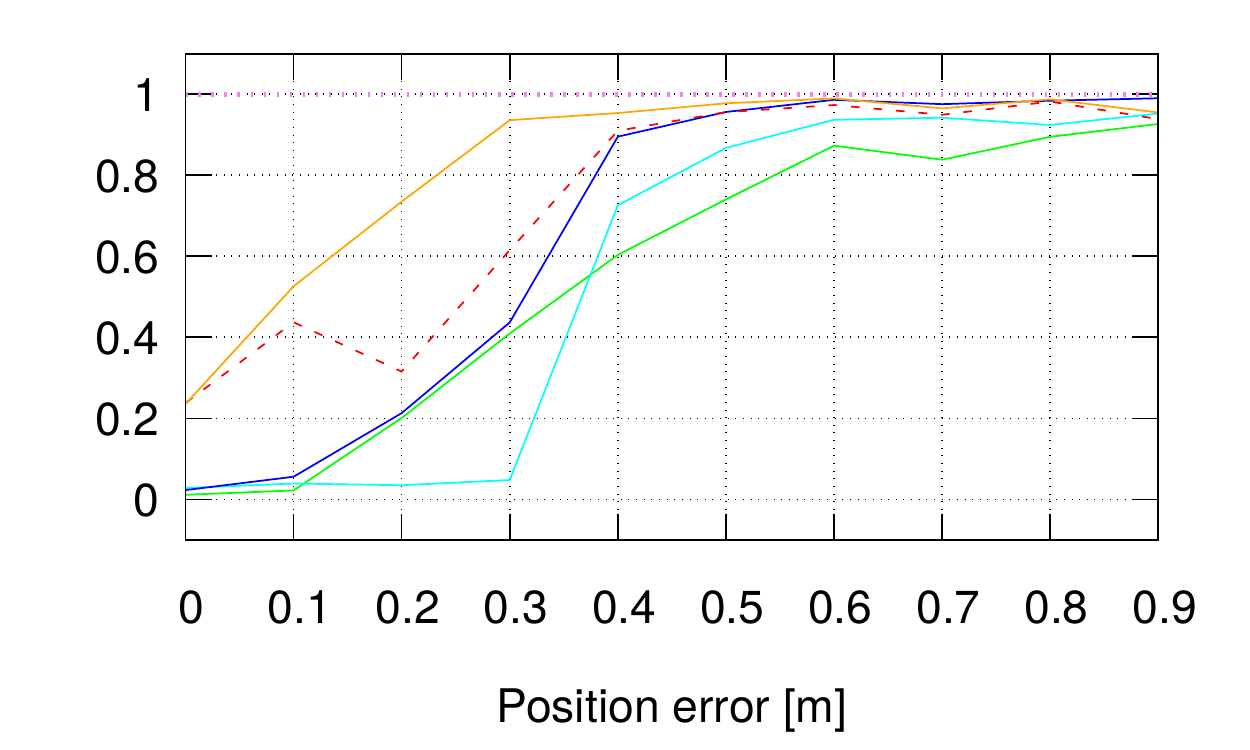}
\caption{Error rate 20\%}
\label{fig:PositionErrorIntersectionRandom20}
\end{subfigure}
\begin{subfigure}[b]{0.32\textwidth}
\includegraphics[clip,trim=0.5cm 0.0cm 0.cm 0.0cm,width=\textwidth]{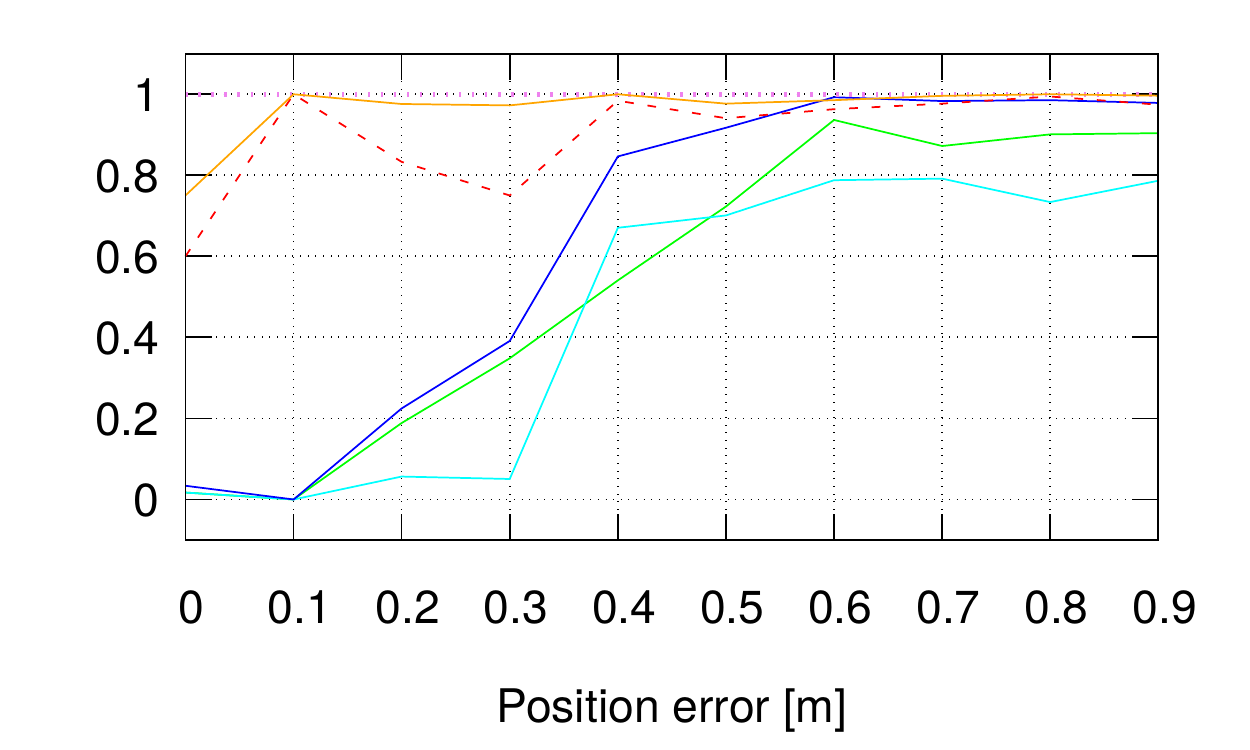}
\caption{Error rate 50\%}
\label{fig:PositionErrorIntersectionRandom50}
\end{subfigure}
\begin{subfigure}[b]{0.32\textwidth}
\includegraphics[clip,trim=0.5cm 0.0cm 0.cm 0.0cm,width=\textwidth]{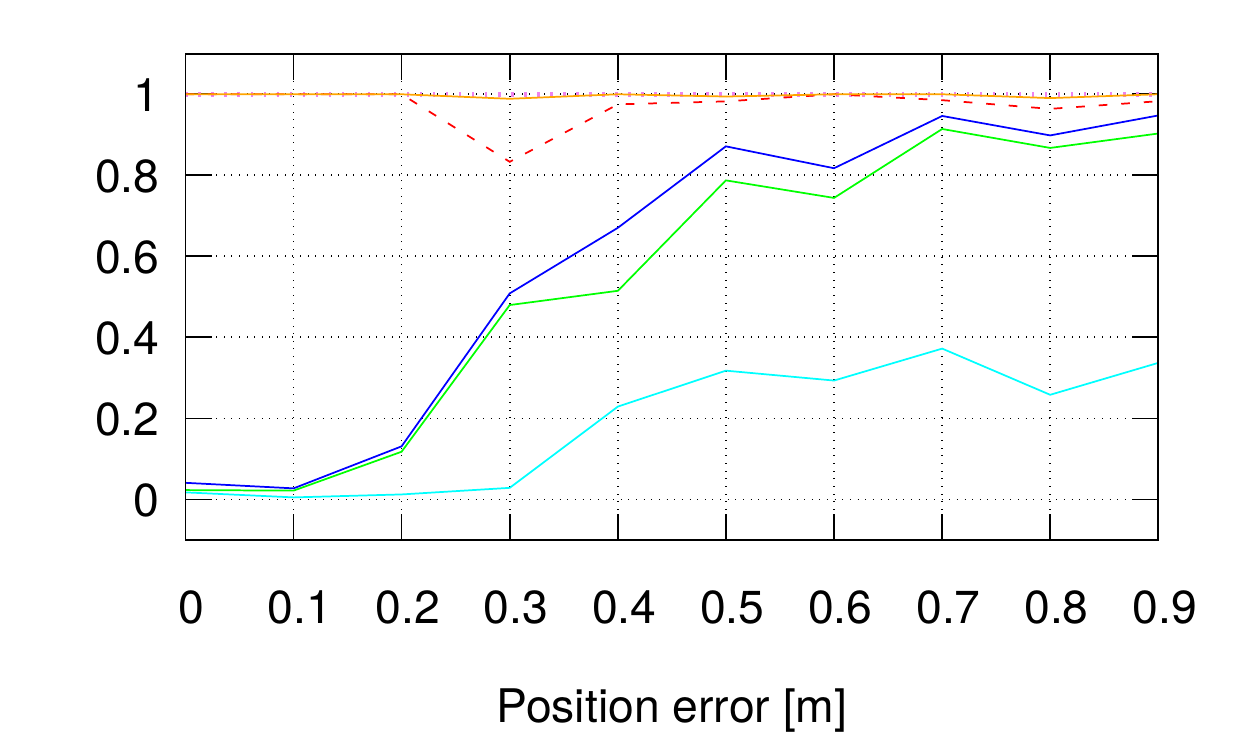}
\caption{Error rate 80\%}
\label{fig:PositionErrorIntersectionRandom80}
\end{subfigure}

\label{fig:PositionErrorRandom}
\caption{Detection result for random position errors injected at different rates. For example, at a $80\%$ error rate an error is injected with a probability of $80\%$ for a given object at a given time step.}
\vspace{-0.5cm}
\end{figure*}

\subsection {Random position error}

We study another realistic error pattern by injecting a position shift error to an individual object based on a fixed error probability (Fig.~\ref{fig:PositionErrorRandom20} - \ref{fig:PositionErrorIntersectionRandom80}).
The sensor checks perform similar to the analysis of the previous section, however, for the intersection scenario the recall of the sensor checks alone reaches only $\approx 90\%$, which can be explained by object occlusions occurring in this scenario. Errors associated with such occluded ground-truth objects cannot be detected by the ego vehicle sensor. 
We thus expect an even higher recall in a real scenario where sensor input is used for the primary instead of the ground truth data. 

The plausibility checks, in contrast to the permanent shifts in Sec.~\ref{sec:EvalPerPosError}, are very well-suited to detect transient position errors, identifying for example in Fig.~\ref{fig:PositionErrorRandom20} transient errors greater than $0.4 \mt$ at a reliable recall rate of $>95\%$. 
The detection capabilities of the plausibility check degrades with a higher error injection probability. This is because the position errors then effectively persist longer (across multiple time frames), resembling more and more the situation of permanent faults studied above.
Overall, the error detection capability of the monitor showcases a slightly better performance in the pedestrian than in the intersection scenario, which is attributed to the more diverse motion patterns of the various object types and the on average higher object speeds in the latter.

\subsection {Velocity errors}

\begin{figure}[h]
\centering
\includegraphics[clip,trim=0.4cm 1.3cm 0.cm 1.5cm,width=0.5\textwidth]{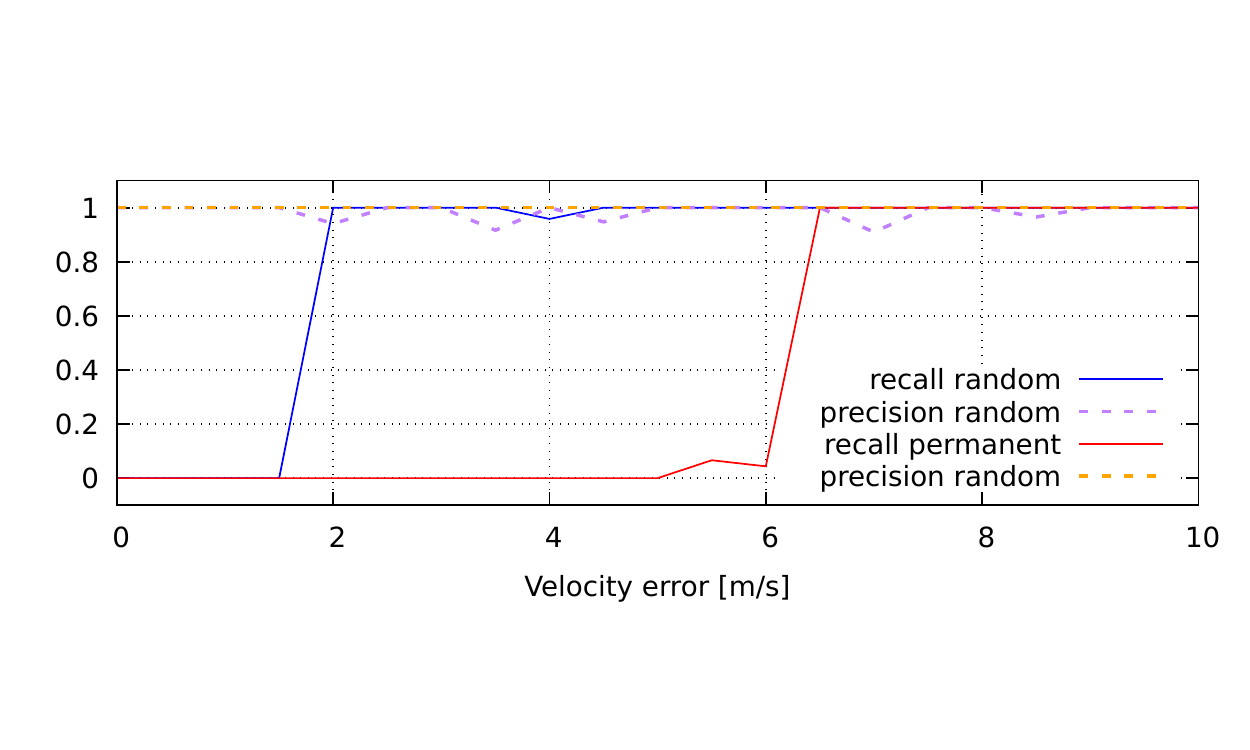}
\caption{Detection capabilities of the plausibility checks for permanent and transient velocity errors.}
\label{fig:velocityError}
\end{figure}

For RSS to work efficiently, it is essential to verify the velocities of objects. They determine for example the expected braking times and safe following distances. 
In order to evaluate the detection capabilities of velocity errors we inject faults to the object speed, but do not affect the direction of the velocity, for simplicity of the analysis.
Fig.~\ref{fig:velocityError} visualizes the results for both permanent and random (transient) speed errors injected at a rate of $10\%$ (sensor update rate $\dt \approx 0.05\scd-0.1\scd$).
As we expected from the theoretical analysis of Sec.~\ref{sec:plaus_theory}, permanent errors can be reliably detected if they are greater than a rather large threshold of $6 \mt/\scd$.
On the other hand, the plausibility checks can efficiently detect already small speed deviations above $2 \mt/\scd$, since the transient errors allow for additional plausibility checking with the help of acceleration limits, see also Fig.~\ref{fig:dv_theo}.

The sensor checks only use position information, and are thus not able to detect any velocity errors.

\subsection{Position error of a real sensor}
Finally, we also explore the error detection capability of our monitor with real \lidar\ data from the NuScenes dataset \cite{nuscenes2019}.
Since no ground truth velocity data is provided here, we restrict ourselves to the analysis of permanent position faults with sensor checks only. As before, the ground truth environment model with injected position errors is used as a primary channel, while the \lidar\ data enables the monitor sensor checks.

\begin{figure}[h]
\vspace{-0.2cm}
\centering

\begin{subfigure}[b]{0.17\textwidth}
\raisebox{0.35cm}{
\includegraphics[clip,trim=0.4cm 0cm 0.cm 0cm,width=\textwidth]{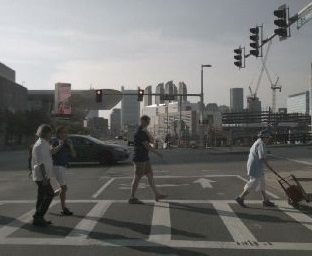}}
\end{subfigure}
\begin{subfigure}[b]{0.3\textwidth}
\includegraphics[width=\textwidth]{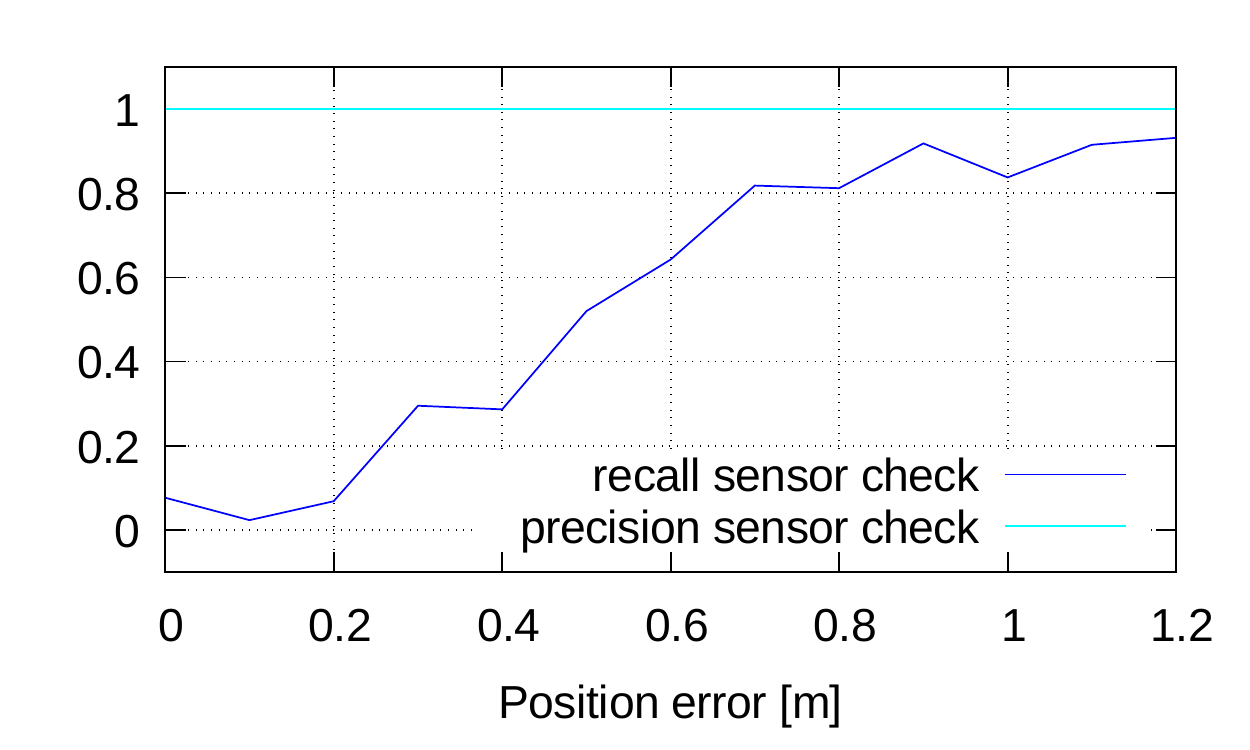}
\end{subfigure}

\caption{Sensor check detection performance for permanent position errors (right) using the NuScenes dataset \cite{nuscenes2019} (left).}
\label{fig:positionNuScenes}
\vspace{-0.2cm}
\end{figure}

Fig. \ref{fig:positionNuScenes} shows that the results are comparable with the results obtained from simulation. The recall of the sensor checks saturates at around $90\%$ for larger position errors, which is again due to partial or full occlusions that the \lidar\ scans fail to detect.

\subsection {Runtime evaluation}
The monitor concept proposed in this article leverages the complementary strengths of sensor and plausibility checks, while remaining functionally simple for high compute efficiency. 
A lightweight implementation is desirable since the checks should ideally be executed on a parallel hardware that is certifiably (e.g. ASIL) robust against soft errors, which typically limits the compute resources.
We here quantify the process latency as a performance metric on selected test hardware, using the two different systems: 
\begin{enumerate}
\item Intel\treg\ Core\ttm\ i9-7900X @ 3.30GHz
\item Intel\treg\ Atom\ttm\ CPU C3934 @ 2.00GHz
\end{enumerate}
The Intel\treg\ Atom\ttm\ CPU represents a potential target platform for ASIL-compliant monitor applications, while the Intel\treg\ Core\ttm\ is a state-of-the-art CPU with higher performance.

\begin{figure}
\centering
\includegraphics[clip,trim=2.cm 5.cm 2.3cm 5.cm,width=0.45\textwidth]{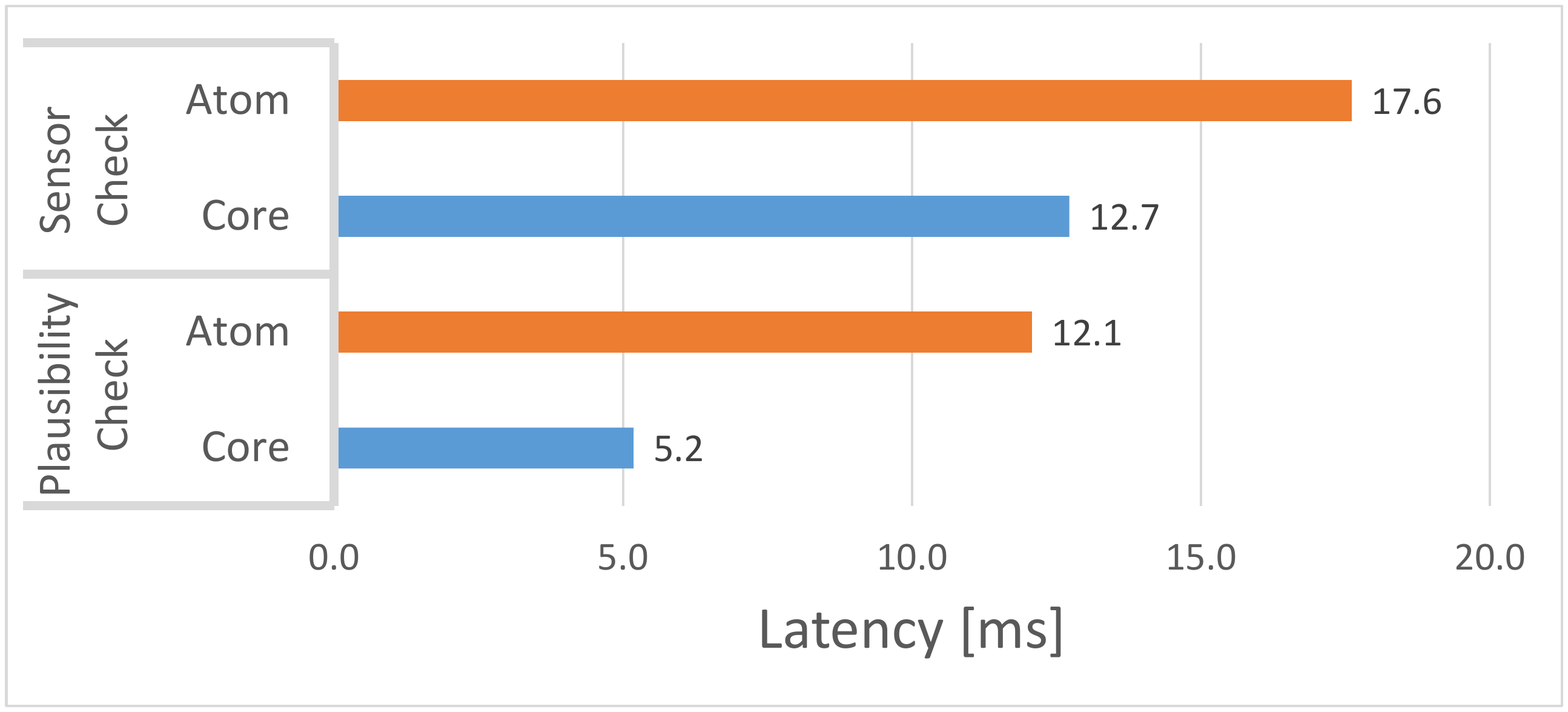}
\caption{Average latency to execute the proposed checks. }
\label{fig:timing}
\vspace{-0.5cm}
\end{figure}

Our tests show (Fig.~\ref{fig:timing}) that the sensor checks require an average of $12\mscd$ on the Core\ttm\ system using the configuration of Table \ref{tab:sim_parameters} (covering a $100\mt \times 100\mt$ area around the ego vehicle, which should be sufficient for urban driving). On the Intel\treg\ Atom\ttm\ the average latency increases to an average $18\mscd$. Plausibility checking here takes an average of $5\mscd$ on the Core\ttm\ system to process about $30$ objects. On the Atom\ttm\ the latency increases to an average $13 \mscd$. 
Those results indicate that our monitor architecture is indeed lightweight, meaning that it is feasible to run the verification process at pseudo-real-time (typically $\approx 50\mscd-100 \mscd$) on a safety-certifiable hardware. Note that the minimum latency target for an unobstructed system execution is eventually determined by the processing time of the parallel primary perception channel. 
The implementation of the code is not yet optimized and runs on a single core. 
The plausibility checks are currently implemented in Python and single-threaded. 
With parallelism and other optimization techniques the latency could therefore be reduced further if required. Also using a non-uniform grid representation \cite{buerkle2020nonuniform} can improve the latency of the sensor checks. 

\section{Conclusion}
\label{sec:conclusion}

We presented a lightweight monitor architecture for automated perception systems, which is realized by a combination of plausibility checks of an object's motion history, and sensor checks that use \lidar\ information. 
To improve overall safety, we envision that this perception monitor is coupled with a supervisor for the AD planning system, such as for example RSS.
Compared to existing monitor approaches, our concept is characterized by two key advantages: i) The checks are designed to be effective yet simple, in order to reduce computational load, ii) We combine diverse sensor-dependent and sensor-independent methods to address both fault-tolerance against common-cause sensor failures and SOTIF.
We evaluated the monitor in simulation, using CARLA, and performed additional tests with the NuScenes dataset. 
Our experiments demonstrate a high recall and precision (both $> 90\%$) for the detection of both permanent and random position errors larger than at most $0.7\mt$.
A comparable detection performance is found for transient speed errors greater than $2\mt/\scd$ or permanent ones greater than $6\mt/\scd$.
Except for the permanent speed errors, those values are more than sufficient to detect the targeted errors of about $1\mt$ position and $3\mt/\scd$ velocity deviations, which we identified as a representative estimate for safety-critical perception errors in a wide spectrum of situations.

We further verified that the proposed monitor is able to run on an ASIL-capable Intel\treg\ Atom\ttm\ CPU with low latency. Explicitly, the sensor check execution required on average $17\mscd$ and the plausibility checks $12\mscd$.
This is a significant reduction of the computational effort compared to our previous work \cite{buerkle2020onlineenv}, where only the creation of the required occupancy grid required more time and cores on an Intel\treg\ Core\ttm\ i9 system.  
For future work, we plan to refine the monitor checks with respect to the safety-relevance of the object attributes and conduct additional test in real world scenarios. 

\section{Acknowledgments}
This research was partially funded by the German Ministry for Economic Affairs and Energy in the project SafeADArchitect (19A20013A) and by the Federal Ministry of Transport and Digital Infrastructure of Germany in the project Providentia++ (01MM19008).
Funding was received from the European Union’s Horizon 2020 research and innovation program within the FOCETA project under grant agreement No. 956123.

\addtolength{\textheight}{-12cm}   



\footnotesize
\bibliographystyle{IEEEtran}
\bibliography{main}



\end{document}